\documentclass[letterpaper,10pt,conference]{ieeeconf}  

\IEEEoverridecommandlockouts                              

\usepackage{epsfig} 
\usepackage{times} 
\usepackage{amsmath} 
\usepackage{amssymb}  
\usepackage{graphicx}
\usepackage{xcolor} 
\usepackage{booktabs}       

\usepackage{amsfonts}
\usepackage{multirow}
\usepackage{chngpage}
\usepackage{soul}
\usepackage[hidelinks]{hyperref}
\usepackage{adjustbox}
\usepackage[flushleft]{threeparttable}

\newcommand{\im}{\mathrm{\mathbf{I}}} 
\newcommand{\dm}{\mathrm{\mathbf{d}}} 
\newcommand{\R}{\mathbb{R}}
\newcommand{\D}{\mathcal{D}}
\newcommand{\loss}{\mathcal{L}}
\newcommand{\real}{\mathrm{r}}
\newcommand{\syn}{\mathrm{s}}
\DeclareMathOperator{\E}{\mathbb{E}}
\newcommand{\fts}{\footnotesize}

\definecolor{Purple}{cmyk}{0.45,0.86,0,0}

\title{\LARGE \bf
	DCL: Differential Contrastive Learning for Geometry-Aware Depth Synthesis
}
\author{
	Yuefan Shen$^{*1}$, Yanchao Yang$^{*\dag2}$, Youyi Zheng$^{\dag1}$, C. Karen Liu$^{2}$, Leonidas J. Guibas$^{2}$
	\thanks{$^{1}$State Key Lab of CAD\&CG, Zhejiang University, Hangzhou, China (email: \{jhonve, youyizheng\}@zju.edu.cn).$^{2}$the Computer Science Department, Stanford University, CA 94305, USA (email: \{yanchaoy, karenliu, guibas\}@cs.stanford.edu). This work was supported by a Hoffman-Yee Research Grant, the NSF grant IIS-1763268, and a Vannevar Bush Faculty fellowship. $^{*}$Equal Contributions, $^{\dag}$Corresponding Authors.}
}

\begin{document}

\maketitle
\thispagestyle{empty}
\pagestyle{empty}

\begin{abstract}
	We describe a method for unpaired realistic depth synthesis that learns diverse variations from the real-world depth scans and ensures geometric consistency between the synthetic and synthesized depth. The synthesized realistic depth can then be used to train task-specific networks facilitating label transfer from the synthetic domain. Unlike existing image synthesis pipelines, where geometries are mostly ignored, we treat geometries carried by the depth scans based on their own existence. We propose differential contrastive learning that explicitly enforces the underlying geometric properties to be invariant regarding the real variations been learned. The resulting depth synthesis method is task-agnostic, and we demonstrate the effectiveness of the proposed synthesis method by extensive evaluations on real-world geometric reasoning tasks. The networks trained with the depth synthesized by our method consistently achieve better performance across a wide range of tasks than state of the art, and can even surpass the networks supervised with full real-world annotations when slightly fine-tuned, showing good transferability. Code is available at: {\color{blue}\url{https://github.com/Jhonve/DCL-DepthSynthesis}}
\end{abstract}

\section{Introduction}

\begin{figure}[t]
	\begin{center}
		\includegraphics[width=1.0\linewidth]{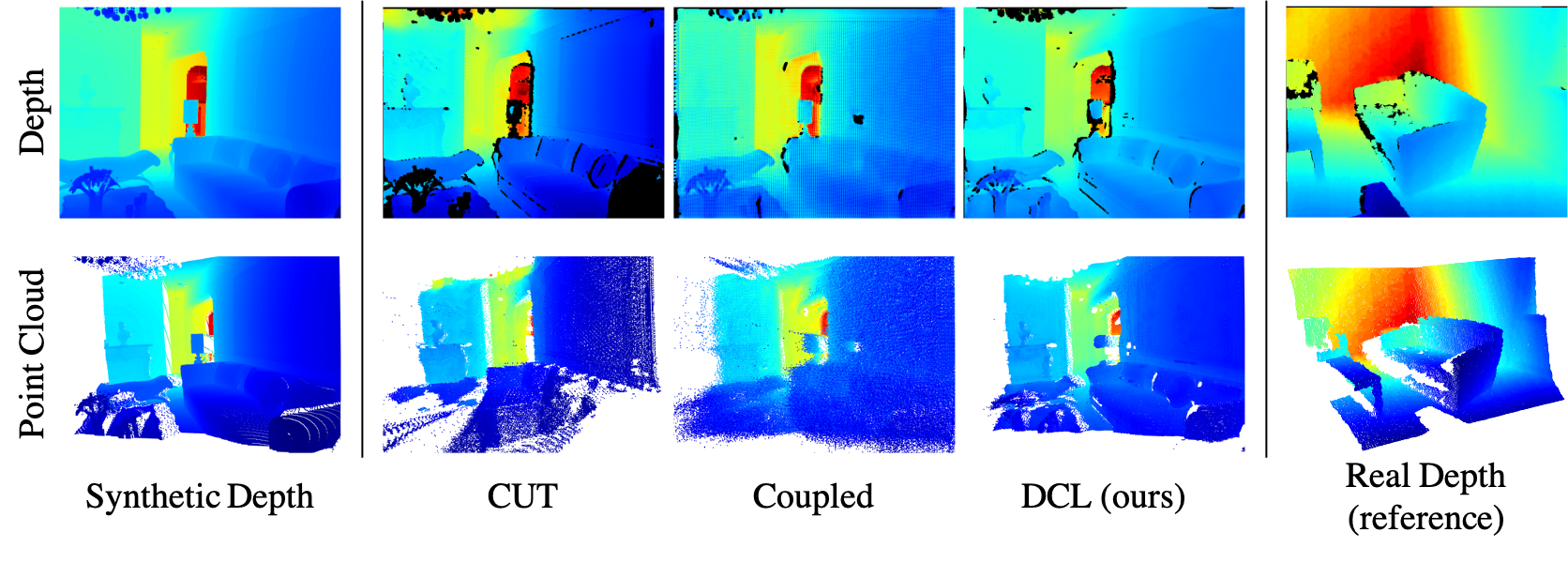}
	\end{center}
	\vspace{-0.6cm}
	\caption{Synthetic to real depth synthesis. Left: clean depth map; Middle: synthesized depth map by CUT \cite{park2020contrastive}, Coupled \cite{gu2020coupled} and DCL (ours); Right: reference real-world depth from ScanNet \cite{dai2017scannet}. Corresponding point clouds are displayed in the second row. Note our method captures missing regions and sensor noise similar to that exhibited in the real-world one. Moreover, our method preserves the underlying geometry as demonstrated in the point clouds, e.g., ours has much fewer out of surface points and distortions.}
	\vspace{-0.8cm}
	\label{fig:teaser}
\end{figure}

Unpaired realistic depth synthesis is important in transferring annotations for geometric reasoning tasks from simulation, where labels can be automatically generated, while label generation in the real world is expensive. 
There exist many works on realistic image synthesis based on generative adversarial networks, yet, there are only a few on realistic depth synthesis.
Traditional methods on realistic depth synthesis either model the real depth variations with an empirical noise model or add random noise and dropout to corrupt the synthetic depth.
Thus, their capability to capture diverse real variations is limited.
On the other hand, learning-based real depth synthesis methods add noise and missing regions to the synthetic depth maps by transformation networks usually trained in an adversarial manner.
Despite the ability to reduce the distributional shift between the synthetic and real domains, the underlying geometric properties are not well preserved and are always subject to undesired distortions, affecting the label transfer efficiency for downstream geometric reasoning tasks (see Fig.~\ref{fig:teaser}).

We treat geometric properties as first-class citizens since depth maps are 2.5D representations of the scene, and the geometric properties carried in depth maps deserve their own existence.
Moreover, we propose differential contrastive learning to learn the real-world depth variations to minimize the distributional shift between domains and explicitly enforce the underlying geometry to be consistent for efficient transfer between domains.
As illustrated in Fig.~\ref{fig:DCL}, the proposed differential contrastive learning first computes differences between features extracted at different spatial locations within each feature map of the depth (the synthetic one and its transformed version).
The resulting differential features are then arranged into positive and negative samples following the terminology used in \cite{chen2020simple,park2020contrastive}.
More explicitly, two differential features computed at the same pair of spatial locations are considered positive; otherwise, those computed at different pairs of spatial locations are considered negative.
Positive and negative samples are then used to compute the InfoNCE loss \cite{oord2018representation}, which serves as the training loss, namely, the differential contrastive loss for learning depth synthesis. 

The motivation of our approach derives from the observation that geometric characteristics can always be captured in the differential forms \cite{spivak1970comprehensive}, e.g., surface normal.
So we explicitly ask the corresponding differential features computed from the depth to be as invariant as possible against the synthesis procedure, which is then ensured by how we select positive and negative samples in the proposed differential contrastive loss.
The resulting unpaired depth synthesis framework can be used for learning realistic variations from any depth sensor of any type, while preserving the geometric properties. 
Moreover, our approach is task agnostic, so the synthesized realistic depth maps, together with synthetic labels, can be used for training any downstream tasks.

We evaluate the quality of the synthesized depth across a broad spectrum of downstream tasks, including depth enhancement, normal estimation, pose estimation, grasping, and semantic segmentation.
The task models trained with the depth map synthesized by our method consistently achieve the best performance when tested on real-world data without fine-tuning.
Our work makes the following contributions: 1) a framework that explicitly models geometric consistency for depth synthesis; 2) a mechanism that prevents geometric distortion by contrasting the feature differences instead of the features themselves; 3) an extensive study of state-of-the-art synthesis methods on a wide range of downstream tasks while achieving top performance.

\section{Related Work}

{\bf Image generation and translation.}
Image generation maps a random noise sampled from a prior distribution to images satisfying a predefined distribution. 
Many works have been proposed to generate diverse and realistic images based on generative adversarial networks (GAN) \cite{goodfellow2014generative}.
Please refer to \cite{pan2019recent} for a detailed overview.
Image translation aims to transform images from one domain to 
another, in either 
paired~\cite{sangkloy2017scribbler} or unpaired~\cite{zhu2017unpaired} settings.
Image translation can help reduce domain gaps \cite{hoffman2018cycada} when it is enforced to preserve task-relevant information \cite{yang2020fda}.
Cycle-consistency is widely used as a regularizer for style transfer \cite{zhu2017unpaired}. 
Recently, contrastive unpaired translation (CUT) \cite{park2020contrastive} proposes contrastive losses on the patch-level features to enforce the similarity of the input and output image features. 
Besides the vast development on image-to-image translation
\cite{choi2018stargan,liu2019few,Gokaslan_2018_ECCV}, little effort has been devoted to depth-to-depth synthesis, where the translation has to align not only the domain noise but also preserve the underlying geometry that is crucial for downstream geometric inferences based on depth.

{\bf Contrastive learning.} 
Based on the InfoNCE loss \cite{oord2018representation}, contrastive learning has been shown effective for self-supervised representation learning \cite{chen2020simple}.
The critical ingredient of contrastive learning is the selection of variations to which we would like the learned representations to be invariant \cite{wang2020understanding}.
Our primary task is not to learn representations that share the invariance of the downstream tasks.
Instead, we learn realistic variations of depth and utilize contrastive learning to take care of geometric properties that should be invariant to the synthesis process.


{\bf Point cloud generation.}
3D point cloud generation is closely related to 2.5D depth map synthesis.
A variational auto-encoder with multi-resolution tree networks is used in \cite{gadelha2018multiresolution} to generate point clouds, and \cite{achlioptas2018learning} studies various GANs and proposes a Gaussian mixture model in the latent space of an autoencoder.
Instead of transforming random noise, \cite{yang2018foldingnet} maps a set of 2D grid points to the target point cloud through deep grid deformation, and \cite{yang2019pointflow} proposes hierarchical modeling of shapes and points using continuous normalizing flows.
At the scene level, \cite{yue2018lidar} generates synthetic point clouds via a virtual lidar in simulation.

{\bf Depth synthesis.}
To inject realistic noise into synthetic depth maps, \cite{landau2015simulating} proposes an empirical noise model of the depth sensor's transmitter/receiver system, which captures sensor-specific noise and may not generalize to different ones.
Similarly, \cite{planche2017depthsynth} explicitly models sensor noise, material properties, and surface geometry for depth synthesis, but is limited to single CAD models.
One can also add random Gaussian noise and dropout to synthesize additive sensor noise and missing regions \cite{seoud2018increasing,litvak2019learning}.
Furthermore, \cite{shrivastava2017learning} applies adversarial training to synthesize realistic hand pose images, and \cite{atapour2019generative} synthesizes holes with a network trained to predict missing regions from RGB images. 
Similarly, \cite{gu2020coupled} learns the hole prediction from real RGBD images, but relies on image translation to bring synthetic images to the real domain such that the learned hole prediction model can be applied on synthetic RGB images.
One can also apply domain adaptation for depth-based predictive tasks \cite{Liu_2019_ICCV,wu2019squeezesegv2}. However, these methods do not generate realistic depth maps.
Our method focuses on realistic depth synthesis, and the synthesized depth maps can be used for any tasks that take depth as input. 
Even though our method can be used in conjunction with domain adaptation methods when specific tasks are known, we treat realistic depth synthesis as our primary goal and evaluate the quality of the synthesized depth maps using specific geometric reasoning tasks.






\section{Method}

\begin{figure}[t]
	\begin{center}
		\vspace{0.2cm}
		\includegraphics[width=1.0\linewidth]{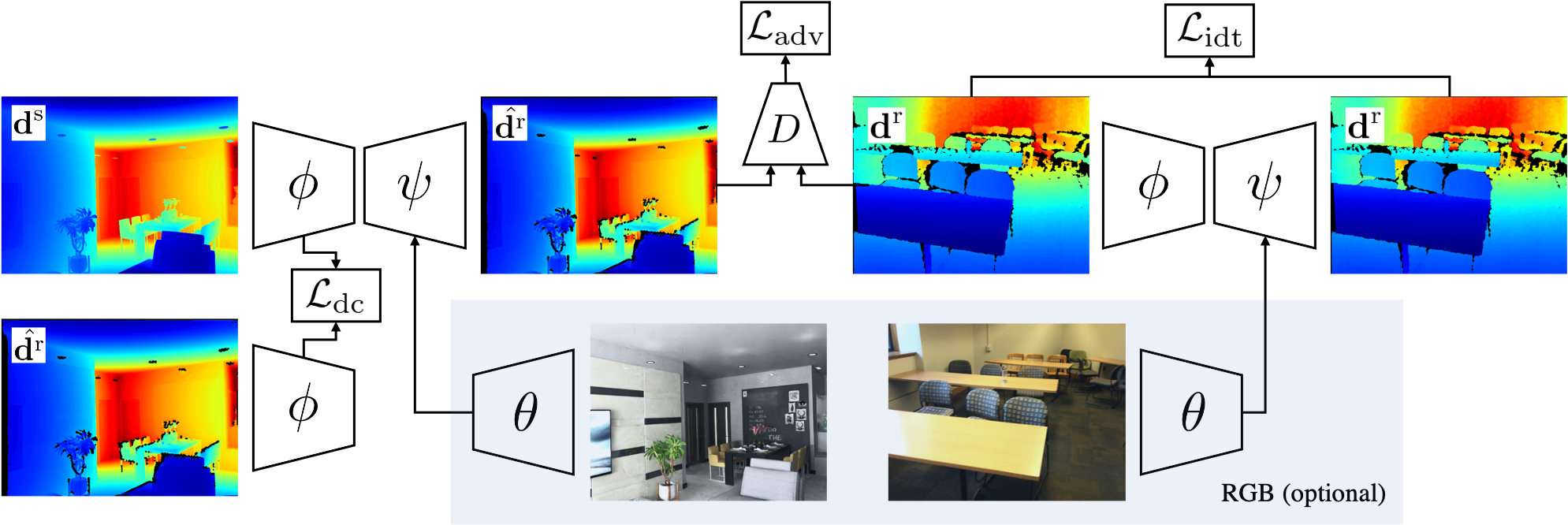}
	\end{center}
	\vspace{-0.5cm}
	\caption{Overview of our method. The depth synthesis network comprises $\phi$ (encoder) and $\psi$ (decoder). Optionally, $\psi$ can take auxiliary information from the aligned RGB image through an image encoder $\theta$. The three loss terms are described in Sec.~\ref{sec:method}.}
	\vspace{-0.6cm}
	\label{fig:pipeline}
\end{figure}

\label{sec:method}

Let $\dm \in \R^{\mathrm{H}\times\mathrm{W}}$ be a depth map, and optionally $\im \in \R^{\mathrm{H}\times\mathrm{W}\times\mathrm{3}}$ be the corresponding color image.
Suppose we have a synthetic (clean) dataset $\D^\syn = \{(\dm^\syn, \im^\syn)\}$ and a real-world (noisy) dataset $\D^\real = \{ (\dm^\real,\im^\real) \}$.
Both of them contain pairs of aligned depth maps and color images. However, there is no pairing between $\D^\syn$ and $\D^\real$, a typical setting of unpaired image synthesis or translation.
Our goal is to learn a mapping between the synthetic depth map $\dm^\syn$ and the real depth map $\dm^\real$, using only unpaired datasets $\D^\syn$ and $\D^\real$.
The overall architecture of our method is illustrated in Fig.~\ref{fig:pipeline}.
Note that the aligned color images are auxiliary, without which our method still runs, and we elaborate in the following.

\subsection{Depth-to-depth Synthesis}

We first illustrate our method using only depth maps for simplicity. We then detail how color images can be incorporated as auxiliary signals to facilitate the synthesis procedure.
Let $\phi$ be an encoder, and $\psi$ be a decoder, together they constitute the transformation network: 
\begin{equation}
	\hat{\dm^\real} = \psi( \phi(\dm^\syn) )
	\label{eq:prediction}
\end{equation}
with $\hat{\dm^\real}$ be the synthesized (noisy) depth map conditioned on the clean depth map $\dm^\syn$.
To enforce statistical similarity between the synthesized depth map $\hat{\dm^\real}$ and the real depth map $\dm^\real$,
we can apply a discriminator network $D$ to minimize the domain discrepancy by adversarial training:
\begin{equation}
	\loss_{\mathrm{adv}} = \E_{\dm^\real\sim \D^\real} \log D(\dm^\real) + \E_{\dm^\syn\sim \D^\syn} \log (1-D(\hat{\dm^\real}))
	\label{eq:loss_adv}
\end{equation}
Since Eq.~\eqref{eq:loss_adv} only helps to reduce the distributional shift, but does not guarantee the consistency of the generated content \cite{taigman2016unsupervised,shrivastava2017learning},
CycleGAN \cite{zhu2017unpaired} resorts to cycle consistency to constrain the transformation.
On the other hand, contrastive unpaired translation (CUT) \cite{park2020contrastive} eliminates the cycle consistency by applying contrastive loss on features from multiple layers of the encoder to preserve the image content. 
Even though it works for image-to-image translation, we observe heavy distortions on the underlying geometric structures of the synthesized depth maps, which hinder the transfer from synthetic domains to real domains.


\begin{figure}[t]
	\begin{center}
		\vspace{0.2cm}
		\includegraphics[width=1.\linewidth]{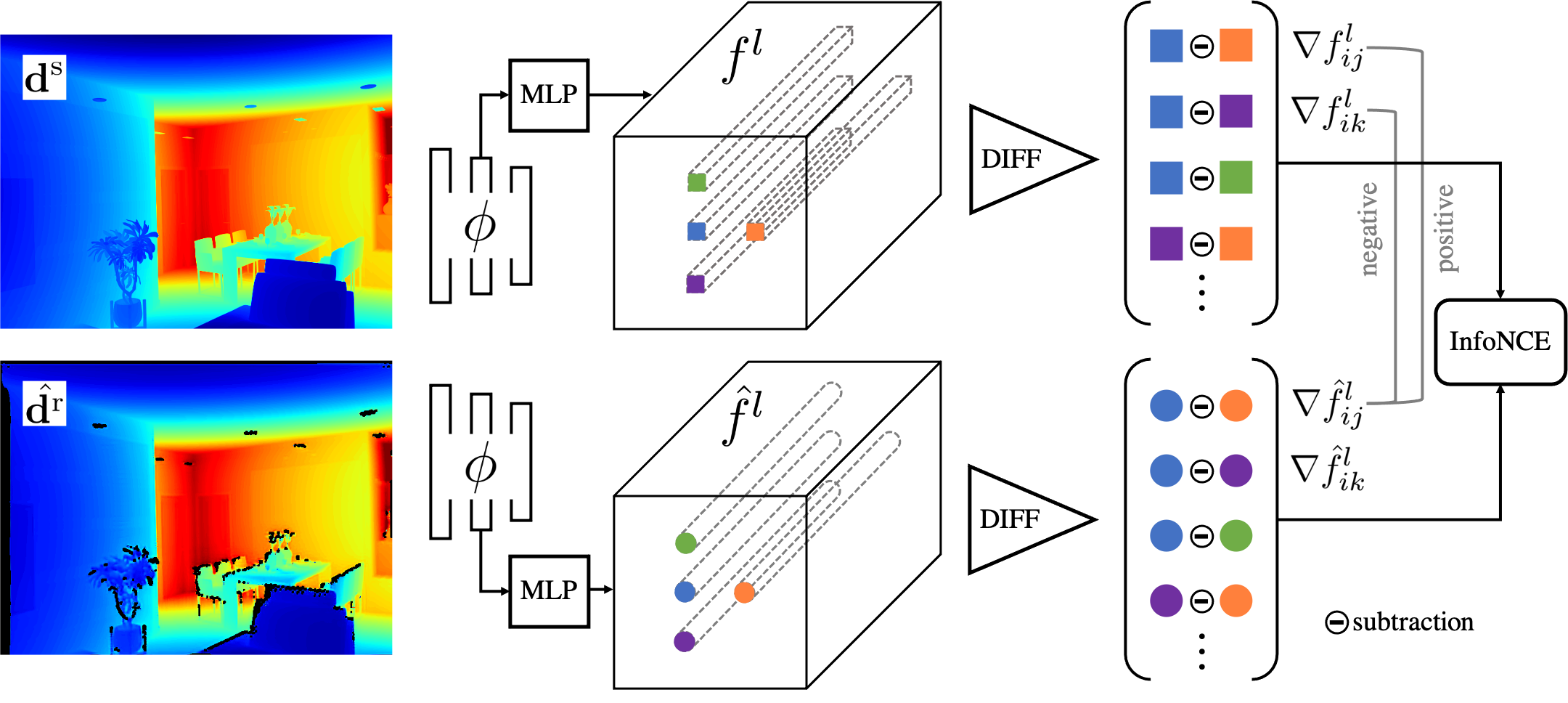}
	\end{center}
	\vspace{-0.5cm}
	\caption{Differential Contrastive Learning. Given a synthetic depth map $\dm^\syn$ and the transformed noisy depth map $\hat{\dm^\real}$, feature maps $f^l$ and $\hat{f^l}$ are extracted from the $l$-th layer of the encoder $\phi$. Differential features are then computed and arranged into positive and negative samples depending on their pairwise spatial locations, e.g., differential features with the same row number are positive samples.}
	\vspace{-0.7cm}
	\label{fig:DCL}
\end{figure}

\subsection{Differential Contrastive Learning}

We aim for a transformation $\psi\circ\phi$ that can capture the complex noise phenomenon in real depth, and, at the same time, preserve the underlying geometry of the clean depth maps for better transfer on geometric reasoning tasks.

Since we ask for geometric invariants of the synthetic depth maps,
we choose to work with the InfoNCE loss \cite{oord2018representation} due to its effectiveness in capturing invariants for self-supervised representation learning \cite{chen2020simple,he2020momentum}.
However, the type of invariants that will be learned with the InfoNCE loss depends mainly on the mechanism to choose positive and negative samples.
For example, in \cite{chen2020simple}, an image and its color distorted version are considered as a pair of positive samples, whereas this same image and another different image are considered as negative samples.
With this sampling strategy, the features learned will be invariant to color distortions but still be discriminative for image identities.

Inspired by the fact that geometric properties can always be captured by their differential forms \cite{spivak1970comprehensive},
we propose to impose an explicit constraint on the underlying geometry of the scene through differential contrastive learning shown in Fig.~\ref{fig:DCL}.
Let $f^l = \phi^l(\dm^\syn)$, $\hat{f}^l = \phi^l(\hat{\dm^\real})$ be the feature maps extracted from the $l$-th layer of the encoder $\phi$ applied on the synthetic depth map $\dm^\syn$ and the synthesized depth map $\hat{\dm^\real}$. Also, let $f^l_i$ be the feature vector from $f^l$ at the spatial location $i$.
We apply the following sampling mechanism to collect positive and negative pairs:
\begin{align}
	\mathbf{positive: \ } & (\nabla\hat{f}^{l}_{ij} = \hat{f}^l_i-\hat{f}^l_j, \nabla f^l_{ij} = f^l_i-f^l_j) \\
	\mathbf{negative: \ } & (\nabla\hat{f}^{l}_{ij} = \hat{f}^l_i-\hat{f}^l_j, \nabla f^l_{ik} = f^l_i-f^l_k)
	\label{eq:sampling}
\end{align}
where $j,k$ are different spatial locations sampled around $i$ following a Gaussian.
Note that each sample consists of two differential vectors (synthetic and synthesized) computed either at the same pair-wise locations (positive) or different pair-wise locations (negative) (see Fig.~\ref{fig:DCL}). Given the InfoNCE loss:
\begin{multline*}
	\loss_{\mathrm{nce}}(\nabla\hat{f}^l_{ij},\nabla f^l_{ij},\{\nabla f^l_{ik}\}_{k\neq j}) = \\
	-\log \dfrac{\exp(\nabla\hat{f}^l_{ij}\cdot\nabla f^l_{ij} / \tau)}{\exp(\nabla\hat{f}^l_{ij}\cdot\nabla f^l_{ij} / \tau) + \sum_k \exp(\nabla\hat{f}^l_{ij}\cdot\nabla f^l_{ik} / \tau)}
\end{multline*}
Our {\bf differential contrastive loss} is defined as:
\begin{equation}
	\loss_{\mathrm{dc}} = \E_{\dm^\syn\sim \D^\syn} \sum_l\sum_{i,j} \loss_{\mathrm{nce}}(\nabla\hat{f}^l_{ij},\nabla f^l_{ij},\{\nabla f^l_{ik}\}_{k\neq j})
	\label{eq:loss_dc}
\end{equation}
here $(i,j,k's)$ can be randomly sampled to avoid enumerating the entire grid. 
The key insight is that, we want the differential features to be similar (invariant) before and after the transformation, i.e., the depth values may be altered due to the noise or missing regions learned from real depth maps; however, the underlying geometric structures captured by the differentials should be similar. 
In other words, the proposed differential contrastive loss explicitly enforces the consistency between geometric structures of the synthetic and synthesized depth maps, while leaving enough flexibility for the transformation network to learn real variations.
Note, differential contrastive losses over feature maps from multiple layers of the encoder $\phi$ are also computed, making it possible to capture both local and global geometric properties.

Given that a global shift in the depth values might be differentiated away and thus can not be detected with the proposed differential contrastive loss,
we apply an identity loss on the real depth map $\dm^\real$ to prevent potential global shifts in the range of the synthesized depth:
\begin{equation}
	\loss_{\mathrm{idt}} = \E_{\dm^\real\sim \D^\real} \|\psi(\phi(\dm^\real)) - \dm^\real\|_1
	\label{eq:loss_idt}
\end{equation}
which is the $L1$ loss between a real depth map and its transformed version. 
The {\bf final training loss} of the proposed differential contrastive learning for synthesizing realistic depth maps from synthetic ones is:
\begin{equation}
	\loss = \loss_{\mathrm{adv}} + \alpha \loss_{\mathrm{dc}} + \beta \loss_{\mathrm{idt}}
	\label{eq:loss_all}
\end{equation}
where $\alpha,\beta$ are the scalar weights, which are set to $1.5$ and $1.0$ for all experiments.




\section{Experiments}

We evaluate the proposed depth synthesis method on multiple downstream tasks, including depth enhancement, normal estimation, pose estimation, grasping, and semantic segmentation. 
Our goal is to have a comprehensive understanding of the capability of our method to learn the noise exhibited in the real-world depth scans by checking the performance of the task-specific networks trained using the synthesized depth on real-world geometric reasoning tasks.

Specifically, the clean synthetic depth maps are first transformed into noisy realistic ones in a task-agnostic manner. We then perform evaluations on each downstream task in three settings: 
1) train a task-specific network using only the labels from the synthetic domain, and test the network directly on a test set from the real domain; 
2) apply task-specific domain adaptation methods, e.g., \cite{hoffman2018cycada,ganin2015unsupervised,you2019universal} using depth from the synthetic and real domains, then test on the real validation set as in 1).
3) fine-tune the previously trained task-specific networks using a small portion of the annotations from the real domain, and then test on the same real test set as in 1).
In the first setting, we like to check how the synthesized depth maps mimic the real ones. In the second, 
we check whether our synthesis helps to reduce the domain gaps when compared to task-specific domain adaptation methods. In the third, we check the usefulness of the weights from the first setting in terms of reducing the number of labels compared to the one supervised with full real annotations.

\subsection{Training Depth Synthesis}

{\bf Datasets.}
For depth enhancement, normal estimation, and semantic segmentation,
we use InteriorNet \cite{li2018interiornet} as the source of synthetic data, and ScanNet \cite{dai2017scannet} as the source of realistic data.
InteriorNet provides depth maps 
rendered from 1.7M interior layouts for different scenes created by professional designers.
We randomly sample 30K depth maps from InteriorNet to form the synthetic dataset. We further split them into a subset of 24K depth maps for training the depth synthesis network and 
the remaining for training task-specific networks.
Each sample in the synthetic dataset consists of a clean depth map and the corresponding annotations for the downstream tasks.
Similarly, we randomly sample 24K real scans from ScanNet for depth synthesis, which contains real-world depth maps 
from 1.5K indoor scenes. 
The raw scans are manually annotated.
For task-specific networks, we follow the dataset filtering in~\cite{jeon2018reconstruction} to guarantee the data quality, after which there are 7K depth maps left. Specifically, 4K depth maps are used for supervised training and testing depth enhancement and normal estimation. In semantic segmentation, we apply all the filtered data for supervised training and testing to ensure reasonable performance. For all tasks, sampled subsets for testing have no overlap regarding frames with the supervised training sets.

For pose estimation and grasping, we use the large-scale GraspNet \cite{fang2020graspnet} as our primary dataset, whose raw scans are used as the real-world depth, and the synthetic depth maps are rendered with the corresponding camera parameters and CAD models.
GraspNet \cite{fang2020graspnet} contains 48.6K depth maps along with annotated object labels, object poses, and ground-truth grasps (center, gripper orientation, and width), which are used for training pose estimation and grasp proposal networks. 
The LineMOD dataset \cite{brachmann2014learning} 
is also used for pose estimation, which contains 20K depth maps and ground-truth object poses.
The detailed split and pre-processing of the data are described in the evaluation sections.

{\bf Baselines.}
A natural depth synthesis baseline is the identity mapping, i.e., the clean depth scans (Baseline).
We also compare with the commonly used empirical noise model \cite{handa2016understanding} (Simulation), and two state-of-the-art image translation methods \cite{zhu2017unpaired} (CycleGAN) and \cite{park2020contrastive} (CUT).
Moreover, we compare to \cite{gu2020coupled} (Coupled), which is the current state-of-the-art on synthetic to real depth synthesis that separately models missing regions and sensor noise based on both depth and color images. 

{\bf Training details.} 
Our network architecture follows that of CUT \cite{park2020contrastive}, which employs an image transformation network consisting of ResNet blocks \cite{he2016deep}.
We extract multi-scale features from five evenly distributed layers of the encoder to compute the differential contrastive loss.
Like CUT, we apply a two-layer MLP with 128 output units and $L2$ normalization on the extracted features before passing them to the loss function.
For the competing methods, aligned color images are always available to the transformation network, except the empirical noise model proposed in \cite{handa2016understanding}, which only relies on geometry to synthesize noise.
We use an Adam optimizer for the training of depth synthesis with differential contrastive loss ($\beta_1=0.5$ and $\beta_2=0.999$) with an initial learning rate of $0.0002$ (same to all downstream tasks).
We set the batch size to 16 and train up to 50 epochs. 
Our method can optionally incorporate RGB images as auxiliary signals during synthesis (DCL w/ rgb). However, 
due to the lack of background textures in GraspNet and LineMOD datasets,  we cannot render corresponding RGB images as auxiliaries, thus we report the depth-only results on them.
Now we detail the evaluations on each downstream task and report the qualitative and quantitative results.

\subsection{Depth Enhancement}

After training the depth synthesis network, we convert the 6K synthetic depth maps to realistic (noisy) ones.
We train a depth enhancement network \cite{jeon2018reconstruction} for each of the synthesized datasets from different synthesis methods.
We also apply the depth enhancement training loss proposed in \cite{jeon2018reconstruction} for all models
, and the training runs for $50$ epochs.

\begin{table}[t]
	\vspace{0.2cm}
	\begin{threeparttable}
		\caption{Depth Enhancement}
		\small
		\centering
		\begin{tabular}{c@{\hspace{3.8ex}}c@{\hspace{3.8ex}}c@{\hspace{3.8ex}}c@{\hspace{3.8ex}}c}
			\toprule
			\fts Method & \fts $\mathrm{RMSE}\downarrow$ & \fts $\mathrm{MAE}\downarrow$ & \fts $\mathrm{PSNR}\uparrow$ & \fts $\mathrm{SSIM}\uparrow$ \\
			\midrule
			\fts Baseline                                 & 0.1871 & 0.0716 & 33.670 & 0.7822\\
			\fts Simulation \cite{handa2016understanding} & 0.1618 & 0.0817 & 36.561 & 0.7955\\
			\fts CycleGAN \cite{zhu2017unpaired}          & 0.1448 & 0.0911 & 38.771 & 0.8469\\
			\fts CUT \cite{park2020contrastive}           & 0.1757 & 0.1186 & 34.870 & 0.8066\\
			\fts Coupled \cite{gu2020coupled}             & 0.1273 & 0.0568 & 41.522 & 0.8524\\
			\fts DCL                                      & 0.1233 & \bf 0.0564 & 42.340 & 0.8692\\
			\fts DCL w/ rgb                               & \bf 0.1198 & 0.0650 & \bf 42.980 & \bf 0.8754\\
			\midrule
			\fts DCL*                                     & 0.0996 & 0.0407 & 47.424 & 0.9096\\
			\fts Supervised 							  & 0.1062 & 0.0437 & 45.426 & 0.9079 \\
			\bottomrule
		\end{tabular}
		\footnotesize{Scores are computed with the networks trained using only the synthesized (noisy) depth maps and the corresponding clean depth maps. Top-performing ones are marked as bold and * means fine-tuning with $10\%$ of the real-world annotations.}
		\label{tab:depth_enhance}
	\end{threeparttable}
	\vspace{-0.7cm}
\end{table}

When the training converges, we test each enhancement network on a preserved real-world test set consists of 300 raw scans and the corresponding clean scans generated from reconstructed meshes.
We report the scores under multiple evaluation metrics: root mean square error (RMSE), mean absolute error (MAE), peak signal-to-noise ratio (PSNR), and structural similarity index measure (SSIM) following \cite{gu2020coupled}.
The first two measure the accuracy of the enhanced depth maps, and the latter two measure the structural similarity of the enhanced depth maps compared to the ground truth.

As shown in Tab.~\ref{tab:depth_enhance}, our method consistently achieves smaller error and higher structural similarity compared to other methods. As expected, the empirical noise model (Simulation) \cite{handa2016understanding} generally performs worse than the learning-based methods.
Note that CUT \cite{park2020contrastive} performs even worse than Simulation\footnote{We have tuned CUT and other competing methods using grid search for their optimal hyper-parameters.}. We conjecture that CUT may capture biased real noise. Hence, its performance is not even as good as the empirical noise model that randomly adds noise without looking at the real scans.
We include the score from a purely supervised model that is trained on a separate training set of 3000 real depth scans to provide a reference on the desired real domain performance.

\begin{figure}[t]
	\begin{center}
	\vspace{0.2cm}
		\includegraphics[width=1.0\linewidth]{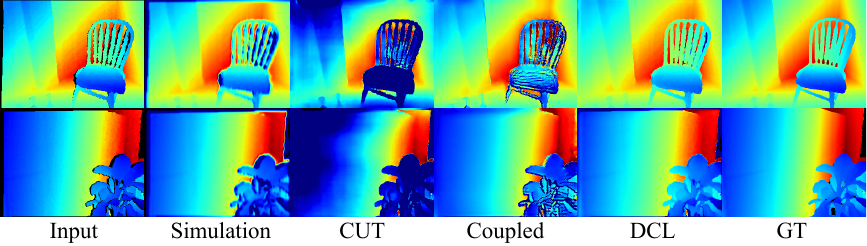}
	\end{center}
	\vspace{-0.5cm}
	\caption{Visual comparison on depth enhancement. From left to right: input real depth scan, Simulation \cite{handa2016understanding}, CUT \cite{park2020contrastive}, Coupled \cite{gu2020coupled}, DCL (ours) and ground-truth.}
	\vspace{-0.6cm}
	\label{fig:enhanceresults}
\end{figure}

We also fine-tune the networks trained with the synthesized depth using $10\%$ of the annotations from the real-world training set that is used to train the supervised baseline in Tab.~\ref{tab:depth_enhance}.
As observed, the model trained with the synthesized depth from DCL surpasses the supervised baseline on all metrics, 
which confirms that the network parameters learned using our synthesized depth maps transfer efficiently to the real world. 
Please see Fig.~\ref{fig:enhanceresults} for visual comparisons. 

\begin{table}[t]
	\vspace{0.2cm}
	\begin{threeparttable}
		\caption{Normal Estimation}
		\small
		\centering
		\begin{tabular}{c@{\hspace{2.4ex}}c@{\hspace{2.4ex}}c@{\hspace{2.4ex}}c@{\hspace{2.4ex}}c@{\hspace{2.4ex}}c}
			\toprule
			\fts Method & \fts $\mathrm{Median}\downarrow$ & \fts $\mathrm{Mean}\downarrow$ & \fts $16\uparrow$ & \fts $22.5\uparrow$ & \fts $30\uparrow$\\
			\midrule
			\fts Baseline                                 & 21.127 & 24.907 & 0.392 & 0.541 & 0.679 \\
			\fts Simulation \cite{handa2016understanding} & 23.725 & 27.473 & 0.349 & 0.485 & 0.623 \\
			\fts CycleGAN \cite{zhu2017unpaired}          & 22.321 & 25.454 & 0.367 & 0.501 & 0.653 \\
			\fts CUT \cite{park2020contrastive}           & 22.492 & 25.082 & 0.361 & 0.507 & 0.665 \\
			\fts Coupled \cite{gu2020coupled}             & 20.049 & 24.629 & 0.420 & 0.557 & 0.677 \\
			\fts DCL                                      & 19.343 & \bf 22.724 & 0.407 & 0.579 & \bf 0.743 \\
			\fts DCL w/ rgb                               & \bf 18.920 & 23.805 & \bf 0.444 & \bf 0.579 & 0.693 \\
			
			\midrule
			\fts Cycada~\cite{hoffman2018cycada}          & 20.694 & 24.763 & 0.402 & 0.550 & 0.681\\
			\fts Cycada~\cite{hoffman2018cycada}+DCL      & \bf 17.120 & \bf 21.567 & \bf 0.460 & \bf 0.649 & \bf 0.787\\
			\midrule
			\fts DCL*                                     & 7.907 & 13.881 & 0.739 & 0.815 & 0.865 \\
			\fts Supervised                               & 8.950 & 14.420 & 0.725 & 0.814 & 0.868 \\
			\bottomrule
		\end{tabular}
		\footnotesize{Normal estimation on real depth scans. Networks are trained with synthesized depth and corresponding normal maps. Top-performing ones are marked as bold and * means fine-tuning with $10\%$ of the annotated real-world data.}
		\vspace{-0.6cm}
		\label{tab:normal}
	\end{threeparttable}
\end{table}

\begin{figure}[t]
	\begin{center}
	\vspace{0.2cm}
		\includegraphics[width=1.0\linewidth]{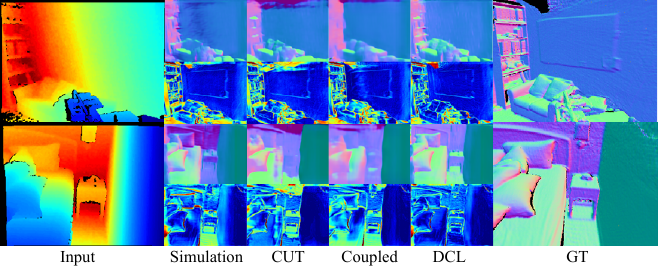}
	\end{center}
	\vspace{-0.5cm}
	\caption{Visual comparison on normal estimation with error maps inserted. Left to right: input real scan, the result of Simulation \cite{handa2016understanding}, CUT \cite{park2020contrastive}, Coupled \cite{gu2020coupled}, DCL (ours) and the ground-truth.}
	\vspace{-0.3cm}
	\label{fig:normalresults}
\end{figure}

\subsection{Surface Normal Estimation}

In this part, we evaluate the quality of the synthesized depth maps by checking the performance of the task-specific networks on normal estimation.
Performance on normal estimation can measure how well the synthesized depth maps preserve the underlying geometry since the surface normal is the cross-product of partial derivatives.
We train the same architecture used for depth enhancement using the L1 loss between predicted normal maps and the ground-truth for 50 epochs. 
We report the angular errors and accuracy.

As shown in Tab.~\ref{tab:normal}, the empirical noise model based method \cite{handa2016understanding} (Simulation) is now consistently worse than the baseline trained with clean depth maps (Baseline), which signals that randomly adding noise could destroy the underlying geometry. 
The same phenomenon is also observed for two other learning-based methods (CycleGAN and CUT).
The normal estimation network trained using the depth synthesized by DCL outperforms the second-best by $3.52\%,7.73\%$ in terms of the median and mean angular errors, respectively.
Involving RGB images in depth synthesis (DCL w/ rgb) has slight improvements for both depth enhancement and normal estimation.
Compared to the domain adaptation method~\cite{hoffman2018cycada}, our method also achieves higher performance. When replacing the source dataset with the synthesized dataset from DCL in Cycada~\cite{hoffman2018cycada}, we can observe a significant improvement, demonstrating the effectiveness of our method.
We further apply fine-tuning on the pre-trained network using $10\%$ of the real-world annotations from the training set of the supervised baseline.
Similarly, the normal estimation network trained using synthesized depth from DCL achieves comparable performance with the supervised baseline. This confirms the quality of the synthesized depth measured by the normal estimation performance directly related to the surface geometry.
Please refer to Fig.~\ref{fig:normalresults} for visual results.

\subsection{Pose Estimation}

\begin{table}[t]
	\vspace{0.2cm}
	\begin{threeparttable}
		\caption{Pose Estimation}
		\small
		\centering
		\begin{tabular}{c@{\hspace{4.8ex}}c@{\hspace{4.8ex}}c@{\hspace{4.8ex}}c@{\hspace{4.8ex}}c}
			\toprule
			&\multicolumn{2}{c}{GraspNet \cite{fang2020graspnet}}&\multicolumn{2}{c}{LineMOD \cite{brachmann2014learning}}\\
			\midrule
			\fts Method & \fts $\mathrm{Acc}\uparrow$ & \fts $\mathrm{Error}\downarrow$ & \fts $\mathrm{Acc}\uparrow$ & \fts $\mathrm{Error}\downarrow$ \\
			\midrule
			\fts Baseline                                 & 57.153 & 55.861 & 44.722 & 62.169 \\
			\fts Simulation \cite{handa2016understanding} & 83.333 & 42.305 & 50.417 & 54.910 \\
			\fts CycleGAN \cite{zhu2017unpaired}          & 78.056 & 43.616 & 63.819 & 51.845 \\
			\fts CUT \cite{park2020contrastive}           & 81.458 & 39.722 & 68.125 & 55.260 \\
			\fts Coupled \cite{gu2020coupled}             & 82.153 & 38.869 & \bf 72.778 & 50.965 \\
			\fts DCL                                      & \bf 87.917 & \bf 36.290 & 71.042 & \bf 42.607 \\
			\midrule
			\fts DANN~\cite{ganin2015unsupervised}        & 92.500 & 32.058 & 86.548 & 43.376 \\
			\fts UAN~\cite{you2019universal}			  & 94.090 & 28.353 & 87.431 & 44.077 \\
			\fts UAN~\cite{you2019universal}+DCL	      & \bf 94.444 & \bf 23.469 & \bf 93.750 & \bf 38.488 \\
			\midrule
			\fts DCL*                                     & 98.264 & 16.906 & 97.986 & 25.087 \\
			\fts Supervised 							  & 99.653 & 16.765 & 99.792 & 23.678 \\
			\bottomrule
		\end{tabular}
		\footnotesize{Results of object pose estimation on GraspNet \cite{fang2020graspnet} and LineMOD \cite{brachmann2014learning}. The prediction error is calculated using the rotation component. The classification accuracy is also reported, and * means fine-tuning with $10\%$ of the annotated real-world data.}
		\label{tab:pose_estimation}
	\end{threeparttable}
	\vspace{-0.8cm}
\end{table}

\begin{figure}[t]
	\begin{center}
		\includegraphics[width=1.0\linewidth]{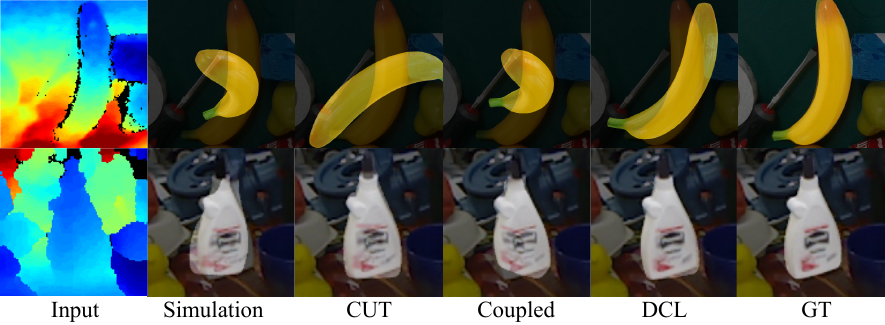}
	\end{center}
	\vspace{-0.5cm}
	\caption{Visual comparison on object pose estimation. Left to right: input real-world scan, object models overlaid on the corresponding color images using the estimated pose. The first row is from GraspNet \cite{fang2020graspnet} and the second is from LineMOD \cite{brachmann2014learning}.}
	\vspace{-0.8cm}
	\label{fig:poseestimation}
\end{figure}

We train task-specific networks that predict poses of objects directly from the depth scans to check how the synthesized depth maps preserve geometric information useful for inferring poses. 
To comply with the evaluation protocol proposed in \cite{brachmann2014learning,bousmalis2017unsupervised}, we crop the scans from GraspNet \cite{fang2020graspnet} using object masks to obtain the input to the pose estimation network.
Similarly, the commonly used object-centric version of LineMOD \cite{brachmann2014learning} is used for further evaluations.
For both datasets, 10K depth scans of 10 randomly chosen objects are preserved for training the pose estimation networks, and the remaining 6K depth scans are used for training the depth synthesis networks.

We use ResNet \cite{he2016deep} as the backbone for all pose estimation networks.
Following the literature \cite{bousmalis2017unsupervised}, the training loss contains a classification error, measured by a cross-entropy loss, and a rotation regression error, measured by an MSE loss.
In addition to the rotation error, which
directly reflects how the fine geometric properties are preserved for pose estimation, we also report the classification accuracy for reference.

The results are shown in Tab.~\ref{tab:pose_estimation}.
As observed, the pose estimation network trained with the synthesized depth from DCL achieves the smallest rotation error on both datasets, e.g., it outperforms the second-best by $6.63\%$ and $16.39\%$ on GraspNet and LineMOD, respectively.
In terms of classification accuracy, our method is comparable with Coupled \cite{gu2020coupled} on the LineMOD dataset despite that Coupled uses aligned RGB images to facilitate the synthesis procedure. Moreover, DCL outperforms Coupled by $7.02\%$ in classification accuracy on the GraspNet dataset.
Overall, DCL is more efficient in learning the real-world variations while maintaining the fine geometries for inferring poses.
Different from dense prediction tasks, here we choose to compare with latent space adversarial domain adaptation methods~ \cite{ganin2015unsupervised,you2019universal}. Again, domain adaptation using synthesized depth from DCL achieves the highest score.
Further, after fine-tuning with $10\%$ of the annotations from the real-world domain, the pose estimation network trained using our synthesized depth achieves similar performance on both datasets compared to the fully supervised baseline.
Please see Fig.~\ref{fig:poseestimation} for visual comparisons.

\subsection{Grasping}

\begin{table}[t]
	\vspace{0.2cm}
	\begin{threeparttable}
		\caption{Grasping}
		\small
		\centering
		\begin{tabular}{c@{\hspace{6.1ex}}c@{\hspace{6.1ex}}c@{\hspace{6.1ex}}c}
			\toprule
			\fts Method & \fts $\mathrm{SR^{Seen}}\uparrow$ & \fts $\mathrm{SR^{Sim}}\uparrow$ & \fts $\mathrm{SR^{Novel}}\uparrow$ \\
			\midrule
			\fts Baseline                                 & 65.0260 & 61.1328 & 56.7448 \\
			\fts Simulation \cite{handa2016understanding} & 80.2866 & 75.0260 & 73.3854 \\
			\fts CycleGAN \cite{zhu2017unpaired}          & 86.6146 & 81.3802 & 77.5391 \\
			\fts CUT \cite{park2020contrastive}           & 92.5000 & 88.5286 & 78.2813 \\
			\fts Coupled \cite{gu2020coupled}             & 83.5938 & 74.4531 & 73.1250 \\
			\fts DCL                                      & \bf 94.2708 & \bf 89.1797 & \bf 82.0833 \\
			\midrule
			\fts Cycada~\cite{hoffman2018cycada}          & 92.3047 & 89.3100 & 82.5651 \\
			\fts Cycada~\cite{hoffman2018cycada}+DCL	  & \bf 95.1823 & \bf 90.7812 & \bf 83.1771 \\
			\midrule
			\fts DCL*                                     & 97.7865 & 95.8594 & 87.0313 \\
			\fts Supervised 							  & 96.1849 & 94.7135 & 84.1016 \\
			\bottomrule
		\end{tabular}
		\footnotesize{Grasping performance on GraspNet \cite{fang2020graspnet}. $\mathrm{SR^{Seen}}, \mathrm{SR^{Sim}}$ and $\mathrm{SR^{Novel}}$ stand for success rates on seen, similar and novel objects, respectively. And * means fine-tuning with $10\%$ of the annotated real-world data.}
		\label{tab:grasping_task}
	\end{threeparttable}
	\vspace{-0.8cm}
\end{table}

\begin{table*}[!t]
	\vspace{0.2cm}
	\begin{threeparttable}
		\caption{Semantic Segmentation}
		\small
		\centering
		\begin{tabular}{c@{\hspace{2.0ex}}c@{\hspace{1.4ex}}c@{\hspace{1.4ex}}c@{\hspace{1.4ex}}c@{\hspace{1.4ex}}c@{\hspace{1ex}}c@{\hspace{1ex}}c@{\hspace{1ex}}c@{\hspace{1ex}}c@{\hspace{1ex}}c@{\hspace{1ex}}c@{\hspace{1ex}}c@{\hspace{1ex}}c@{\hspace{1ex}}c@{\hspace{1ex}}c@{\hspace{1.7ex}}c@{\hspace{1ex}}c}
			\toprule
			\fts Method &\fts \rotatebox[origin=c]{55}{wall} &\fts \rotatebox[origin=c]{55}{floor} &\fts \rotatebox[origin=c]{55}{cabinet} & \fts\rotatebox[origin=c]{55}{bed} & \fts\rotatebox[origin=c]{55}{chair} & \fts\rotatebox[origin=c]{55}{sofa} &\fts \rotatebox[origin=c]{55}{table} &\fts \rotatebox[origin=c]{55}{door} &\fts \rotatebox[origin=c]{55}{window} &\fts \rotatebox[origin=c]{55}{desk} &\fts \rotatebox[origin=c]{55}{curtain} &\fts \rotatebox[origin=c]{55}{ceiling} &\fts \rotatebox[origin=c]{55}{fridge} &\fts \rotatebox[origin=c]{55}{tv} &\fts \rotatebox[origin=c]{55}{others} &\fts \rotatebox[origin=c]{55}{mIoU-15} &\fts \rotatebox[origin=c]{55}{mIoU-12} \\
			\midrule
			\fts Baseline                                 & 39.52 & 41.90 & 2.96 & 0.97 & 0.04 & 5.45 & 1.97 & 1.25 & 2.50 & 0.0 & 2.90 & 4.75 & 0.0 & 0.0 & 9.78 & 7.60 & 9.50 \\
			\fts Simulation \cite{handa2016understanding} & 38.50 & 47.67 & 10.47 & 22.40 & 8.88 & 10.30 & 9.49 & 2.31 & 4.77 & 3.05 & 3.97 & 28.72 & 0.03 & 0.0 & 13.63 & 13.61 & 16.27 \\
			\fts CycleGAN \cite{zhu2017unpaired}          & 39.11 & 47.47 & 10.39 & 21.48 & 6.49 & 9.14 & 9.99 & 1.56 & 4.22 & 2.47 & 4.37 & 30.12 & 0.04 & 0.0 & 14.31 & 13.41 & 16.22 \\
			\fts CUT \cite{park2020contrastive}           & 45.23 & 60.49 & 12.84 & 13.03 & 4.34 & 9.37 & 5.72 & 4.55 & 6.76 & 1.66 & 8.43 & 22.46 & 0.83 & 0.84 & 11.98 & 13.90 & 16.88 \\
			\fts Coupled \cite{gu2020coupled}             & 38.85 & 25.47 & 8.52 & 13.57 & 3.33 & 2.28 & 1.02 & 4.44 & 2.99 & 0.84 & 8.56 & 21.11 & 0.14 & 0.09 & 11.15 & 9.49 & 11.57 \\
			\fts DCL                                      & 45.25 & 61.35 & 11.60 & 17.67 & 8.20 & 11.10 & 7.14 & 7.72 & 8.13 & 5.15 & 6.92 & 19.49 & 1.81 & 3.43 & 11.83 & \bf 15.12 & \bf 17.78 \\
			\fts DCL w/ rgb                      & 46.11 & 67.53 & 12.48 & 10.04 & 8.16 & 11.65 & 4.52 & 3.16 & 4.68 & 2.81 & 8.72 & 26.67 & 1.06 & 0.13 & 15.30 & 14.74 & 17.64\\
			\midrule
			\fts Cycada~\cite{hoffman2018cycada}                          & 54.81 & 72.97 & 13.18 & 26.05 & 9.76 & 12.45 & 13.45 & 8.80 & 10.33 & 7.85 & 7.29 & 39.63 & 5.77 & 4.89 & 16.36 & 20.02 & 23.60 \\
			\fts Cycada~\cite{hoffman2018cycada}+DCL                      & 58.62 & 74.84 & 12.83 & 32.11 & 10.04 & 16.15 & 12.83 & 11.81 & 10.35 & 6.94 & 10.85 & 32.93 & 5.54 & 4.56 & 15.83 & \bf 21.08 & \bf 24.67 \\
			\midrule
			\fts DCL*                                     & 75.62 & 87.56 & 49.50 & 69.84 & 58.01 & 59.81 & 60.91 & 43.22 & 51.78 & 42.09 & 47.90 & 83.53 & 38.51 & 44.75 & 42.15 & 57.01 & 59.49 \\
			\fts Supervised                               & 72.43 & 88.45 & 52.57 & 70.97 & 54.15 & 55.60 & 52.83 & 36.30 & 34.31 & 43.25 & 37.66 & 81.64 & 22.65 & 56.18 & 39.25 & 53.22 & 55.44 \\
			\bottomrule
		\end{tabular}
		\footnotesize{Results on semantic segmentation. We report the intersection-over-union (IoU) scores for each class, and the mean intersection-over-union (mIoU) scores over subsets of the semantic classes (mIoU-15, mIoU-12). The network trained with the depth synthesized by DCL performs well on most classes and consistently outperforms the others in terms of mIoUs. * means fine-tuning with $25\%$ of the real-world annotations.}
		\vspace{-0.6cm}
		\label{tab:semseg}
	\end{threeparttable}
\end{table*}

Grasping also relies on an accurate understanding of the geometries from the depth scans to enable object manipulations. We first apply the depth synthesis methods on the training subset from GraspNet \cite{fang2020graspnet}, which contains 25.6K depth maps.
Then we train GG-CNN \cite{Morrison2018closing} on the synthesized depth to predict the ground-truth grasp proposals. 
To evaluate, we choose the three most confident grasps from the predicted proposals following \cite{Morrison2018closing}.
A grasp is considered successful if its intersection-over-union with the ground truth is larger than 0.5. 
Furthermore, we report the success rate on three different testing subsets of GraspNet, e.g., the test sets for objects seen during training, objects similar to the training objects, and novel objects.

As shown in Tab.~\ref{tab:grasping_task}, the network trained using the depth scans synthesized by DCL achieves the highest success rate among the competing methods.
It outperforms the second-best by $1.91\%$, $0.74\%$, and $4.86\%$ on the test split of seen, similar, and novel objects, respectively.
This shows that DCL learns the realistic variations exhibited in the real-world depth scans for seen objects and generalizes better by preventing potential overfitting to distorted geometries in the synthesized depth. 
Similarly, when the grasp proposal network trained using the depth synthesized by DCL is fine-tuned with $10\%$ of the real-world annotations, it outperforms the supervised baseline trained with full annotations. Moreover, compared with directly applying Cycda~\cite{hoffman2018cycada}, the synthesized depth scans from DCL help achieve better performance.
The results on grasping confirm again that the geometry which enables physical manipulations of the objects is well preserved by our method.

\subsection{Semantic Segmentation}

We train DeepLabv3+ \cite{chen2018encoder} for 50 epochs with the cross-entropy loss for semantic segmentation.
Due to the class mismatch between InteriorNet and ScanNet, we choose a common set of fifteen classes for training since our goal is to validate the synthesis networks' performance for learning realistic sensor noise but not to adapt for class imbalance.
We also report the performance of a purely supervised model trained with 6000 manually annotated real depth scans from ScanNet.

We use the intersection-over-union (IoU) score to measure the quality of the predicted semantic masks. 
We also report the mean intersection-over-union (mIoU) scores across different subsets of classes in Tab.~\ref{tab:semseg}.
Coupled \cite{gu2020coupled} now lags compared to other learning-based methods. Our method still achieves the best performance on most of the classes and the mean IoUs.
However, due to the imbalanced class distributions, we can still observe a gap compared to the supervised baseline.

\begin{figure}[!t]
	\begin{center}
		\includegraphics[width=1.0\linewidth]{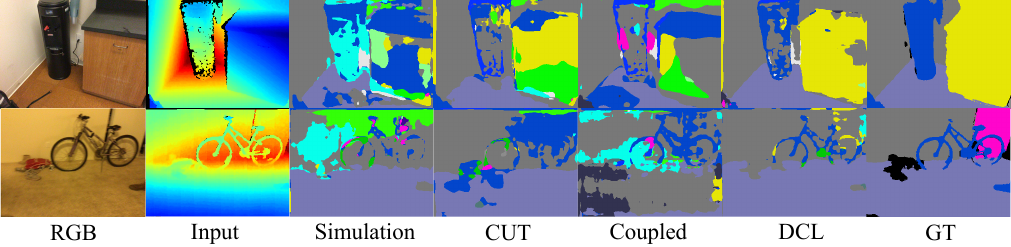}
	\end{center}
	\vspace{-0.4cm}
	\caption{Visual comparison on semantic segmentation from real depth. Left to right: the reference RGB image, input real scan, the result of Simulation \cite{handa2016understanding}, CUT \cite{park2020contrastive}, Coupled \cite{gu2020coupled}, DCL (ours), and the ground-truth.}
	\vspace{-0.3cm}
	\label{fig:segresults}
\end{figure}

To check how the pre-trained weights help reduce the number of real-world annotations needed for semantic segmentation, like previous tasks, we also fine-tune the pre-trained network with $25\%$ of the annotations of the supervised baseline. 
We report the scores in Tab.~\ref{tab:semseg}. 
As observed, the network pre-trained with the depth synthesized by our method now surpasses the supervised baseline. 
Moreover, we find that the fine-tuned model has higher scores in classes that appear more frequently in the synthetic domain, e.g., window and curtain.
Please see Fig.~\ref{fig:segresults} for visual comparisons. Again, the depth scans synthesized by DCL help achieve better domain adaptation performance than Cycada~\cite{hoffman2018cycada}.

\subsection{Analysis and Ablation Studies}
\begin{figure}[t]
	\begin{center}
		\includegraphics[width=1.0\linewidth]{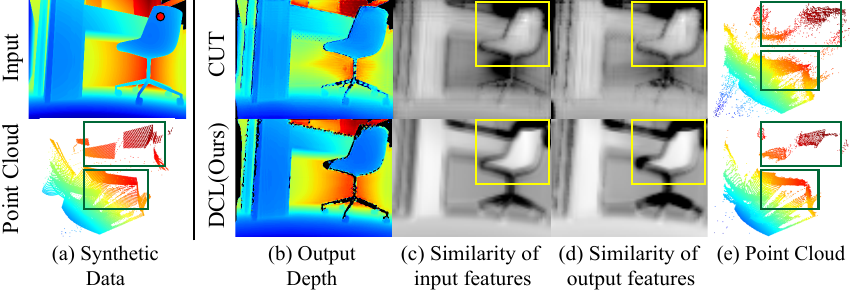}
	\end{center}
	\vspace{-0.4cm}
	\caption{Visualization of the similarities between learned features from DCL and CUT \cite{park2020contrastive}. Given a query point shown as a red dot in the input depth ((a) top) and the output depth maps ((b), top: CUT, bottom: DCL), we visualize the feature similarity to the query point feature both for the features extracted from the input depth (c) and the output depth (d). We also show point cloud visualizations from the top view to examine the geometric distortions (e).}
	\vspace{-0.7cm}
	\label{fig:simvis}
\end{figure}

To further understand how our proposed DCL affects the feature learning in the synthesis procedure, 
we study the feature space from the middle layer of $\phi$. 
As shown in Fig.~\ref{fig:simvis}, 
given an input and output depth map, 
we calculate the similarities between feature maps and a query point feature in the input feature map. 
One can see that features learned with CUT~\cite{park2020contrastive} tend to be dissimilar in different positions even they have the same underlying geometries. 
However, with the proposed DCL, features have higher similarities where the underlying geometries are similar. 
This confirms that DCL prevents undesired distortions by not directly pushing away the features in the differential contrastive learning process.

We perform ablation studies on each loss term in Eq.~\eqref{eq:loss_all} with the depth enhancement task. 
We train the depth synthesis network using five different combinations of the coefficients $\alpha$ (weight on the differential contrastive loss) and $\beta$ (weight on the identity loss). After training the depth synthesis network, we train a depth enhancement network using the synthesized depth maps from each depth synthesis network. Specific coefficients and their corresponding scores are reported in Tab.~\ref{tab:ablation}. These results confirm that the two terms in our proposed loss Eq.~\eqref{eq:loss_all} for depth synthesis are necessary and effective.

\begin{table}[t]
	\vspace{0.2cm}
	\begin{threeparttable}
		\caption{Ablation Studies}
		\centering
			\begin{tabular}{cccccc}
				\toprule
				row $\#$ & $\alpha$ \ \ \ $\beta$  & $\mathrm{RMSE\downarrow}$ & $\mathrm{MAE\downarrow}$ & $\mathrm{PSNR\uparrow}$ & $\mathrm{SSIM\uparrow}$ \\
				\midrule
				(a) & 1.5\ \ \ 0.0 & 0.1813 & 0.0701 & 34.31 & 0.7888\\
				(b) & 0.0\ \ \ 1.0 & 0.2624 & 0.2104 & 26.97 & 0.7181\\
				(c) & 1.0\ \ \ 1.0 & 0.1213 & 0.0653 & 42.423 & \bf0.8767\\
				(d) & 2.0\ \ \ 1.0 & 0.1399 & 0.0664 & 39.65 & 0.8441\\
				(e) & 1.5\ \ \ 1.0 & \bf0.1198 & \bf0.0650 & \bf42.98 & 0.8754\\
				\bottomrule
			\end{tabular}
		\footnotesize{Ablation study on the effectiveness of each term in Eq.~\eqref{eq:loss_all}. 
Evaluations for different sets of the coefficient are performed on the depth enhancement task with the same metrics as in Tab.~\ref{tab:depth_enhance}}
		\label{tab:ablation}
	\end{threeparttable}
	\vspace{-0.7cm}
\end{table}

\section{Discussion}

We have proposed an effective synthetic-to-real depth synthesis method based on differential contrastive learning, which enforces the differential features to be invariant through the synthesis procedure.
Extensive evaluations on downstream geometric reasoning tasks indicate that our method learns diverse variations from the real-world scans and preserves the geometric information crucial for real-world applications.
Our method is end-to-end trainable and does not need to deal with hole generation and value degeneration separately.
However, since our method is GAN-based, it currently does not guarantee to synthesize material-specific realistic noise, which may need further evaluations to show the impact.
In the future, we would like to extend our method to higher-order differential features.
Moreover, since our synthesis method is task-agnostic, we like to see it be used with other task-specific domain adaptation techniques to further improve real-world performance.






{\small
	\bibliographystyle{IEEEtran}
	\bibliography{IEEEabrv}
}

\newpage

\section*{\textbf{Appendix}}

We have proposed an effective synthetic-to-real depth synthesis method based on differential contrastive learning, which enforces the differential features to be invariant through the synthesis procedure.
Extensive evaluations on downstream geometric reasoning tasks indicate that our method learns diverse variations from the real-world scans and preserves the geometric information crucial for real-world applications.
Our method is end-to-end trainable and does not need to deal with hole generation and value degeneration separately.
However, since our method is GAN-based, it currently does not guarantee to synthesize material-specific realistic noise, which may need further evaluations to show the impact.
In the future, we would like to extend our method to higher-order differential features.
Moreover, since our synthesis method is task-agnostic, we like to see it be used with other task-specific domain adaptation techniques to further improve real-world performance.

\subsection*{A. Visual Illustration of Differential Contrastive Learning}

\begin{figure}[ht]
	\begin{center}
		\includegraphics[width=0.93\linewidth]{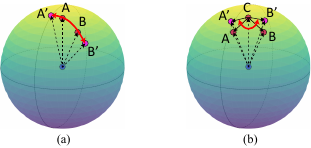}
	\end{center}
	\caption{Illustration of how contrastive learning on features (a) and the proposed differentials work. In (a), features at different locations are pushed away, disregarding their geometries. However, in the proposed differential contrastive learning, the differentials at different pairs of locations are pushed away while the feature themselves can remain close if necessary.}
	\label{fig:sphere}
\end{figure}

Here, we illustrate the advantage of the proposed differential constrastive learning in detail.
Since DCL is performed in the feature space,
where the inputs in the primal space are now embedded into a high-dimensional manifold, 
the differentials thus capture the geometry of the manifold induced by the encoder.
To be more explicit, we can consider the differentials as a measure of the consistency between features, which we illustrate in the following example.

Suppose there are locations, A, B (and C) on the input depth or depth feature map, and the same locations on the output depth or feature map are A', B' (please refer to Fig.~\ref{fig:sphere}).

A method that applies contrastive learning as in CUT \cite{park2020contrastive} will require the features $F_A, F_{A'}$ to be the same and $F_B, F_{B'}$ to be the same, however, $F_A, F_B$ to be different.
This would be problematic since A, B could be two nearby points, and their geometric properties should be similar(Fig.~\ref{fig:sphere} (a)).
Next, we explain how DCL can help relieve this case (Fig.~\ref{fig:sphere} (b)).

As mentioned, DCL requires the differentials to be dis-similar, i.e., $F_A - F_C$ will be dis-similar to $F_{B}-F_{C}$, 
in this case, 
all the three features $F_A, F_B, F_C$ can still be similar, as long as their differentials are different (Fig.~\ref{fig:sphere} (b)).

\subsection*{B. Analysis on Ablation Studies}
From Tab.~\ref{tab:ablation} in our main paper, we can make the following observations:
\begin{itemize}
	\item (a) v.s. (b): the differential contrastive loss is much more effective than the identity loss. Note both (a) and (b) have the optimal coefficient for the activated term. The differential contrastive loss (a) outperforms the identity loss (b) by $30.9\%$, $66.7\%$, in terms of RMSE, and MAE respectively. This confirms that differential contrastive loss is effective in preserving the underlying geometric properties of the depth maps.
	\item (b) v.s. (c): when the identity loss is activated, increasing the weight on the differential contrastive term will significantly increase the performance, which confirms again the effectiveness of the differential contrastive term when the identity loss term is non-zero.
	\item (c) v.s. (d): extremely large weight on the differential contrastive term will be counterproductive. For example, increasing $\alpha$ from 1.0 to 2.0 while fixing $\beta$ to 1.0 will actually decrease the performance by $15.3\%$, $38.9\%$ in terms of RMSE, and MAE respectively. The possible reason is that extremely large $\alpha$ may render the synthesis network too stiff for learning diverse variations in the real scans.
	\item (c), (d), (e): this confirms that $\alpha=1.5$ and $\beta=1.0$ are the optimal weights for our loss in Eq.~(\ref{eq:loss_all}).
	\item (e) v.s. (a): when the weight on the identity loss term is decreased to zero, while fixing the weight on the differential contrastive loss, the performance also drops (but not as big as the drop caused by putting $\alpha$ to zero, i.e., (e) v.s. (b)), which confirms the necessity of the identity loss for preventing possible global range shift.
\end{itemize}
These observations about our ablation studies confirm that the two terms in our proposed loss (Eq.~(\ref{eq:loss_all})) for depth synthesis are necessary and effective.

\subsection*{C. The Shift in the Distribution of Classes between ScanNet and InteriorNet}

\begin{figure}[ht]
	\begin{center}
		\includegraphics[width=1.0\linewidth]{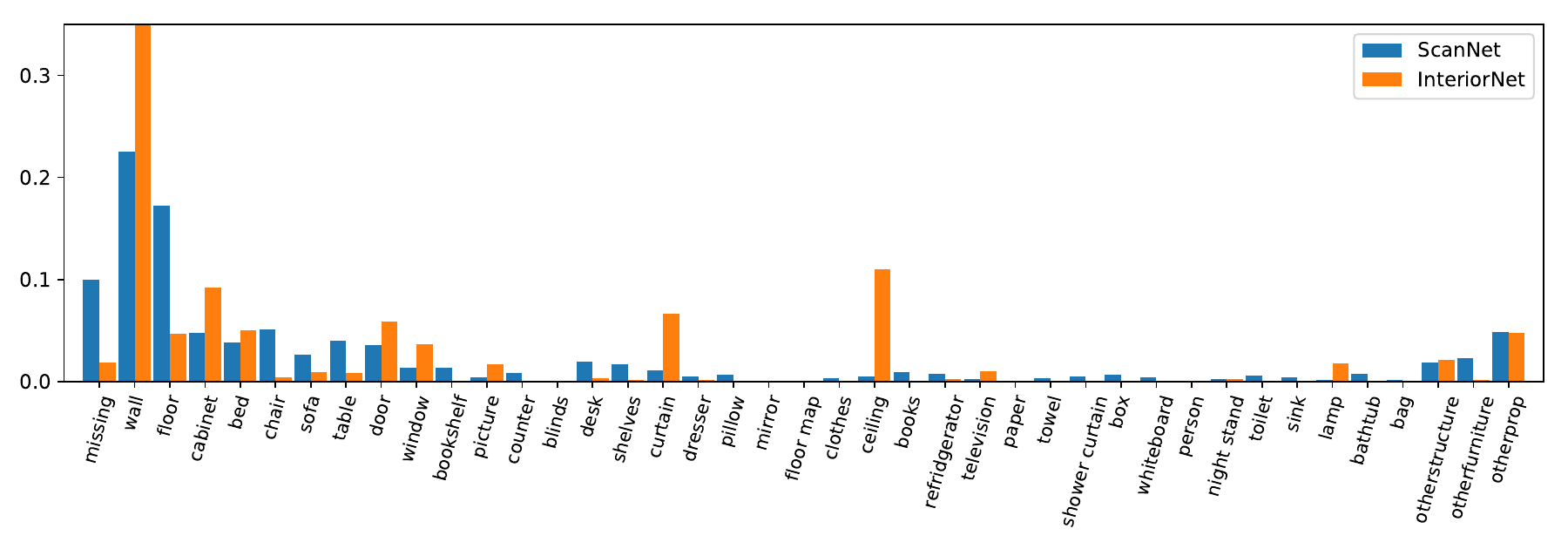}
	\end{center}
	\caption{Distributions of the semantic classes. Blue: ScanNet; Orange: InteriorNet. One can see a big difference between these two distributions. This will complicate the domain gap and make it harder to analyze the effectiveness of the marginal distributions' alignment. Thus we select the most common classes for evaluations.}
	\label{fig:classdistrib}
\end{figure}

\begin{table*}[t]
	\centering
	\begin{adjustbox}{max width = \textwidth}
		\begin{threeparttable}
			\caption{FID Scores}
			\centering
			\begin{tabular}{ccccc}
				\toprule
				& \multicolumn{4}{c}{FID $\downarrow$}\\ \cline{2-5} 
				Method & apple to orange  & horse to zebra & summer to winter & Monet to photo \\
				\midrule
				CycleGAN \cite{zhu2017unpaired} & \bf173.9479 & 118.6636 & 127.1005 & 173.6303 \\
				CUT \cite{park2020contrastive} & 180.4313 & \bf63.11354 & 144.3511 & 163.8327 \\
				DCL (ours) & 174.1408 & 110.9439 & \bf86.92452 & \bf145.7468 \\
				\bottomrule
			\end{tabular}
			{Quantitative evaluation of the style transfer on color images. Scores are the Frechet Inception Distance (FID). Note, our method is competitive compared to the other two state-of-the-art methods that are designed for color image translation.}
			\label{tab:fid}
		\end{threeparttable}
	\end{adjustbox}
	\vspace{-0.4cm}
\end{table*}

We plot the distributions of the semantic classes for both ScanNet~\cite{dai2017scannet} and InteriorNet~\cite{li2018interiornet} in Fig.~\ref{fig:classdistrib}.
As observed, there is a big difference between these two distributions, which will impose another layer of complexity on the domain shift between ScanNet and InteriorNet, i.e., besides the domain shift caused by realistic noise, we now have a content shift that can not be solved by solely adding back the real noise.
To analyze the effectiveness of the depth synthesis methods in a more direct manner, we select the most common classes between the two for evaluations.

\subsection*{D. More Visual Results}

\begin{figure}[ht]
	\begin{center}
		\includegraphics[width=1.0\linewidth]{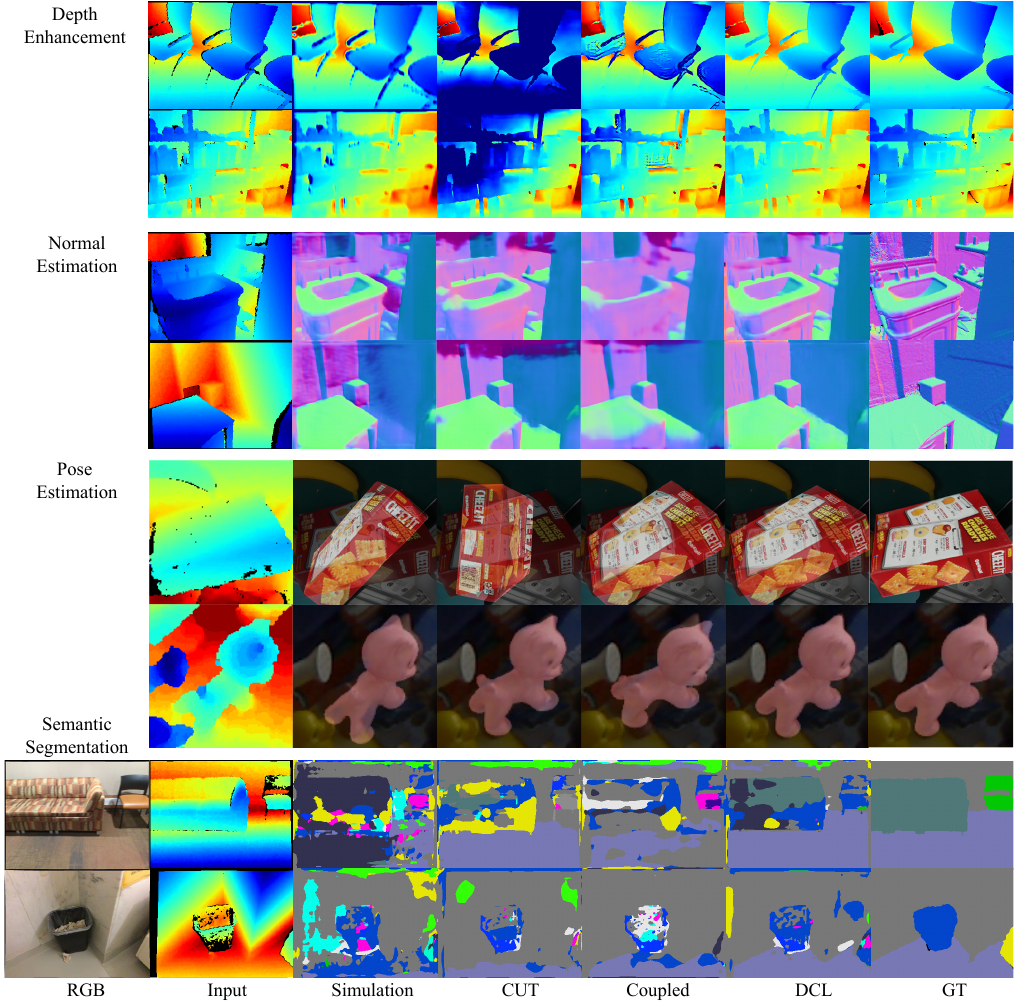}
	\end{center}
	\caption{Visual comparison on four different downstream tasks. From left to right: (the reference RGB image, ) input real scan depth, the result of Simulation~\cite{handa2016understanding}, CUT~\cite{park2020contrastive}, Coupled~\cite{gu2020coupled}, DCL (ours), and the ground-truth.}
	\label{fig:more_vis}
\end{figure}

Here, we show more visual results of our different downstream tasks in Fig.~\ref{fig:more_vis}.

\subsection*{E. More Synthesized Depth}
\begin{figure}[ht]
	\vspace{0.0cm}
	\begin{center}
		\includegraphics[width=1.0\linewidth]{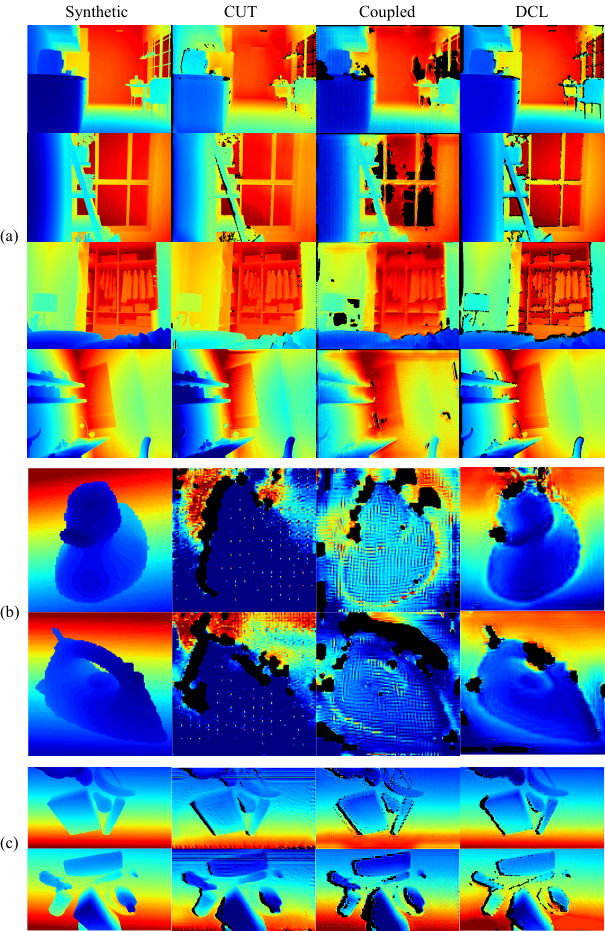}
	\end{center}
	\caption{More visual results on depth synthesis of three datasets ((a) InteriorNet~\cite{li2018interiornet}, (b) LineMOD~\cite{brachmann2014learning}, (c) GraspNet~\cite{fang2020graspnet}). Compared to CUT~\cite{park2020contrastive} and Coupled~\cite{gu2020coupled}.}
	\label{fig:synthesized}
\end{figure}

We include more synthesized realistic depth maps in Fig.~\ref{fig:synthesized}.

\subsection*{F. RGB Image Translation}

\begin{figure*}[t]
	\vspace{-12.0cm}
	\begin{center}
		\includegraphics[width=1.0\linewidth]{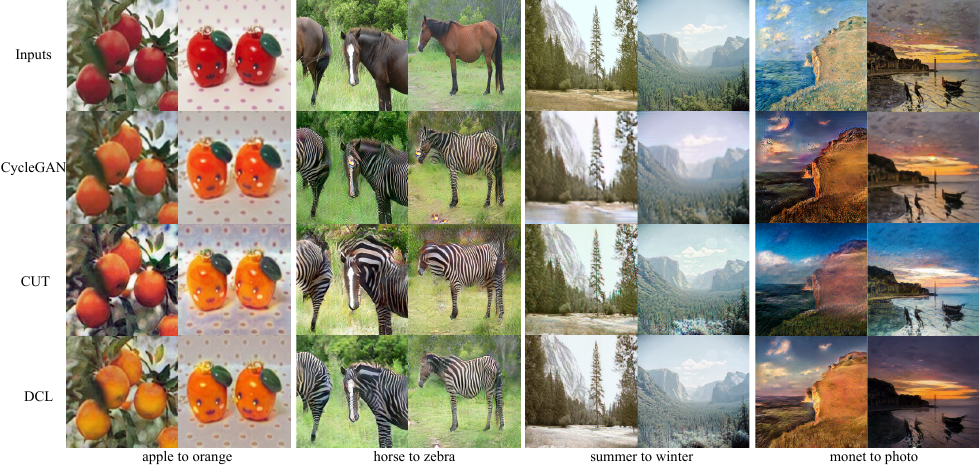}
	\end{center}
	\caption{RGB image translation. We also apply our differential contrastive learning for depth synthesis on the color image translation task and compare with two other state-of-the-art image translation methods CycleGAN \cite{zhu2017unpaired} and CUT \cite{park2020contrastive}. Even though our method is designed to preserve the underlying geometric structures in the depth maps, one can also observe that our method can maintain the image structure and, in some cases, performs even better than the other methods which are designed for color image translation. For example, in the horse to zebra case, our method generates fewer artifacts around the zebra's neck.}
	\vspace{-0.0cm}
	\label{fig:imagetask}
\end{figure*}

Our method is designed to preserve the underlying geometric properties in the synthesized depth maps.
However, it can be directly applied to image-to-image translation without any modification.
For readers' interest, we train our model on image-to-image translation and compare to the other state-of-the-art in Fig.~\ref{fig:imagetask} and Tab.~\ref{tab:fid}.
One can see that our method is comparable to the other image translation methods.
Moreover, our method can preserve the underlying image structure better in some cases even though it is designed to maintain the geometries (e.g., horse to zebra, fewer artifacts around the neck of the zebra).

\end{document}